%% file: main.tex
\documentclass{article}

    \PassOptionsToPackage{numbers, compress}{natbib}

\usepackage[final]{neurips_2022}

\usepackage[final]{neurips_2022}

\usepackage[utf8]{inputenc} 
\usepackage[T1]{fontenc}    
\usepackage{hyperref}       
\usepackage{url}            
\usepackage{booktabs}       
\usepackage{amsfonts}       
\usepackage{nicefrac}       
\usepackage{microtype}      
\usepackage{xcolor}         
\usepackage{amsmath}
\usepackage{amssymb}
\usepackage{adjustbox}
\usepackage{tabularx}
\usepackage{wrapfig}
\usepackage{tikz}
\usetikzlibrary{calc}

\usepackage{xparse}
\usepackage{dsfont}
\usepackage{csquotes}
\usepackage{mathtools}
\usepackage{placeins}
\usepackage{caption}
\usepackage{pgfplotstable}

\usepackage[ruled, vlined]{algorithm2e}

\SetCommentSty{mycommfont}

\input{notation}

\title{CascadeXML: Rethinking Transformers for End-to-end Multi-resolution Training in Extreme Multi-label Classification}

\author{Siddhant Kharbanda \quad 
Atmadeep Banerjee \quad Erik Schultheis \quad Rohit Babbar \\
Department of Computer Science, Aalto University, Finland \\
\{firstname.lastname\}@aalto.fi 
}

\begin{document}

\maketitle

\input{Abstract}
\input{Intro}
\input{Motivation}
\input{Background}
\input{Method}
\input{Experiments}
\input{Discussion}
\input{Related}
\input{Conclusion}

\bibliographystyle{abbrv}
\bibliography{lib}

\clearpage
\appendix
\input{AppendixA.tex}
\input{AppendixB.tex}

\end{document}

%% file: notation.tex
\NewDocumentCommand{\reals}{}{\mathds{R}}

\NewDocumentCommand{\indsymb}{}{\mathds{1}}
\DeclareMathOperator{\expectsymb}{\mathds{E}}

\DeclareMathOperator{\probsymb}{\mathds{P}}

\newcommand{\featurespace}{\mathcal{X}}

\newcommand{\labelset}{\mathsf{y}}
\newcommand{\numlabels}{L}
\newcommand{\numlevels}{T}
\newcommand{\nummeta}{K}

\NewDocumentCommand{\wrapbrackets}{m}{\left[ #1 \right]}

\NewDocumentCommand{\expectation}{e{_}o}{
    \def\decorated{\expectsymb\IfNoValueTF{#1}{}{_{#1}}}
    \IfNoValueTF{#2}{\decorated}{\decorated\!\wrapbrackets{#2}}
}

\NewDocumentCommand{\probability}{o}{
    \IfNoValueTF{#1}{\probsymb}{\probsymb\!\wrapbrackets{#1}}
}

\NewDocumentCommand{\indicator}{o}{
    \IfNoValueTF{#1}{\indsymb}{\indsymb\!\wrapbrackets{#1}}
}

\NewDocumentCommand{\intrange}{o}{
    \wrapbrackets{#1}
}

\NewDocumentCommand{\set}{m}{
    \left\{#1\right\}
}

\NewDocumentCommand{\encoder}{o}{
    \IfNoValueTF{#1}{\Phi}{\Phi_{\text{#1}}}
}

\NewDocumentCommand{\labelfeature}{o}{
    \IfNoValueTF{#1}{z}{z^{(#1)}}
}
\newcommand{\instance}{\mathbf{x}}
\newcommand{\labelvec}{\mathbf{y}}
\newcommand{\clustersymb}{\mathfrak{C}}
\newcommand{\resolutionsymb}{\mathfrak{R}}
\newcommand{\hlt}{\mathfrak{H}}
\newcommand{\shortlistsymb}{\mathcal{S}}
\NewDocumentCommand{\levelemb}{m}{
    \boldsymbol\vartheta^{(#1)}
}
\NewDocumentCommand{\cluster}{mo}{
    \IfNoValueTF{#2}{\clustersymb^{(#1)}}{\clustersymb^{(#1)}_{#2}}
}
\NewDocumentCommand{\resolution}{m}{
    \resolutionsymb^{(#1)}
}
\NewDocumentCommand{\refinecluster}{m}{
    \resolutionsymb^{(#1)}_{\downarrow}
}
\NewDocumentCommand{\coarsencluster}{m}{
    \resolutionsymb^{(#1)}_{\uparrow}
}
\NewDocumentCommand{\shortlist}{m}{
    \shortlistsymb^{(#1)}
}
\NewDocumentCommand{\labelsatlevel}{mm}{
    \resolutionsymb^{(#1)}\!\left(#2\right)
}

\makeatletter
\newcommand{\raisemath}[1]{\mathpalette{\raisem@th{#1}}}
\newcommand{\raisem@th}[3]{\raisebox{#1}{$#2#3$}}
\makeatother

%% file: Abstract.tex
\begin{abstract}
Extreme Multi-label Text Classification (XMC) involves learning a classifier that can assign an input with a subset of most relevant labels from millions of label choices.
Recent approaches, such as XR-Transformer and LightXML, leverage a transformer instance to achieve state-of-the-art performance. 
However, in this process, these approaches need to make various trade-offs between performance and computational requirements.
A major shortcoming, as compared to the Bi-LSTM based AttentionXML, is that they fail to keep separate feature representations for each resolution in a label tree. 
We thus propose CascadeXML, an end-to-end multi-resolution learning pipeline, which can harness the multi-layered architecture of a transformer model for attending to different label resolutions with separate feature representations. 
CascadeXML significantly outperforms all existing approaches with non-trivial gains obtained on benchmark datasets consisting of up to three million labels. 
Code for CascadeXML will be made publicly available at \url{https://github.com/xmc-aalto/cascadexml}.
\end{abstract}

%% file: Intro.tex
\section{Introduction}

Extreme multi-label text classification (XMC) deals with the problem of predicting the most relevant subset of labels from an enormously large label space spanning up to millions of labels. 
Over the years, extreme classification has found many applications in e-commerce, like product recommendation \cite{chang2021extreme} and dynamic search advertisement \cite{Jain19}, document tagging \cite{Agrawal2013MultilabelLW} and open-domain question answering \cite{Chang2020Pre-training, lee-etal-2019-latent} and thus, practical solutions to this problem can have significant and far-reaching impact.

The output space in extreme classification is not only large but also, the distribution of instances among labels follows Zipf's law such that a large fraction of them are \emph{tail labels} \cite{Babbar17, jain2016extreme}. For example, in a dataset of Wikipedia corpus containing 500K labels, only 2.1\% labels annotate more than 100 data points and 60\% of the labels annotate less than 10 data points! 
This inherently makes learning meaningful representations at an extreme scale a challenging problem. 
Further, as the memory and computational requirements grow linearly with the size of the label space, naive treatment via classical methods and off-the-shelf solvers fails to deal with such large output spaces \cite{Chang20, Yen16}. 

In extreme scenarios with millions of labels, a \emph{surrogate} step of shortlisting candidate labels becomes crucial to perform extreme classification and has been adapted by many popular approaches \cite{Dahiya21, Jiang2021, kharbanda2021embedding, Ye20, zhang2021fast}.
For each data point, these approaches first solve a simpler task with coarse label-clusters or \emph{meta-labels} as the label space to create a shortlist of relevant labels for the extreme classification task.
This effectively reduces the training time and computational complexity of the extreme task to that of the shortlisting step which is $\mathcal{O}(\sqrt{L})$, where $L$ is the number of labels.

Existing works leverage label representation as a centroid of their annotated instances to create a Hierarchical Label Tree (HLT) \cite{Parabel, you18}. 
While earlier tree-based methods like Parabel \cite{Parabel} and Bonsai \cite{Khandagale19} used the entire HLT for extreme classification, 
many recent works \cite{Chang20, Dahiya21, Jiang2021, kharbanda2021embedding, Ye20} use label clusters only at a certain level of the HLT as meta-labels which are in turn used to shortlist candidate labels for the extreme task. 
In contrast, XR-Transformer \cite{zhang2021fast} and AttentionXML \cite{you18} leverage multiple levels of the HLT, such that each level corresponds to a certain label \textit{resolution} in the tree structure \cite{zhang2021fast}.

%% file: Motivation.tex
\subsection{Motivation: Strengths and Weaknesses of Current Approaches}

The earliest work to successfully combine label-tree based shortlisting and attention-based deep encoders is
AttentionXML \cite{you18}, which employs multiple Bi-LSTMs models that are trained sequentially for each tree
resolution. More recent approaches \cite{Chang20, Jiang2021, Ye20, zhang2021fast} replaced the model architecture with a
more powerful transformer model \cite{transformer} and fine-tune a pre-trained instance such as Bert \cite{Bert}. 
Without careful designs, such models are computationally very expensive and earlier works \cite{Chang20, Ye20} could not effectively leverage transformers for both computation and performance on XMC tasks.
Next, we discuss two contemporary works LightXML \cite{Jiang2021} and XR-Transformer \cite{zhang2021fast}.

\begin{table}[t]
    \centering
  \begin{tabular}{lcccc}
    \toprule
    & AttentionXML & LightXML    & XR-Transformer & CascadeXML  \\ \midrule
    Base Model         & Bi-LSTM      & Transformer & Transformer   & Transformer \\
    Attention Maps & adaptive   & shared    & shared      & adaptive  \\
    Single Encoder       & $\times$     & \checkmark  & \checkmark    & \checkmark  \\
    Multi-level HLT & \checkmark   & $\times$    & \checkmark    & \checkmark  \\
    End-to-End         & $\times$     & \checkmark  & $\times$      & \checkmark  \\
    Full Resolution   & \checkmark   & \checkmark  & $\times$      & \checkmark  \\
    \bottomrule
  \end{tabular}
  \vspace{1em}
  \caption{Strengths and limitations of current XMC methods: LightXML makes end-to-end training possible, but only admits a single level of the HLT. 
XR-Transformer allows for multiple levels, but its iterative feature learning does not train the transformer encoder at the highest resolution at all \cite[Table 6]{zhang2021fast}. 
However, by changing from multiple Bi-LSTMs to a single transformer, both LightXML and XR-Transformer lost AttentionXML's ability to adapt attention maps to each resolution.}
  \label{tab:xmc-methods}
\end{table}

\textbf{LightXML} introduces the concept of dynamic negative sampling, which replaces pre-computed label shortlists with a dynamically calculated shortlist that changes as the model's weights get updated. 
This enables end-to-end training with a single model by using the final feature representation of the transformer encoder for both the meta- and the extreme-classification task. 
Unfortunately, these two tasks appear to interfere with one another \cite{kharbanda2021embedding}. We hypothesize that this is because the meta task needs the attention maps to focus on different tokens than the extreme task.
Furthermore, it only uses a single-level tree, which prevents scaling to the largest datasets.

\textbf{XR-Transformer} takes inspiration from multi-resolution approaches in computer vision like super resolution \cite{lai2017deep} and progressive growing of GANs \cite{karras2017progressive, karras2019style}, and enables multiple resolutions through iterative training. 
However, unlike progressively grown GANs, which predict only at the highest resolution, XR-Transformer needs predictions across all resolutions for its progressive shortlisting pipeline, but uses representations trained at a single resolution. 
In practice, this leads to XR-Transformer having a complex multi-stage pipeline where the transformer model is iteratively trained up to a certain resolution and then frozen.
This is followed by a \emph{re-clustering} and \emph{re-training} of multiple classifiers, working at different resolutions, with the same \emph{fixed} transformer features \cite[Alg. 2]{zhang2021fast}. 
For datasets with over 500K labels, the transformer is only trained on up to $2^{15}$ clusters \cite[Table 6]{zhang2021fast}, and the resulting features are used for the extreme task.

Unlike AttentionXML, using multiple instances of transformer models becomes undesirable due to their computational overhead. 
This enforces LightXML and XR-Transformer to make different trade-offs when leveraging a single transformer model for XMC tasks compared to AttentionXML, see \autoref{tab:xmc-methods}. 
In this paper, we present a method, \emph{CascadeXML}, that combines the strengths of these approaches, thus creating an end to-end trainable multi-resolution learning pipeline which trains a single transformer model across multiple resolutions in a way that allows the creation of label resolution specific attention maps and feature embeddings.

\subsection{CascadeXML in a Nutshell}
\label{sec:nutshell}
The key insights that enable calculating resolution-specific representations with a single forward pass through the
transformer are \textbf{a)} that the shortlisting tasks get easier for lower label resolution and 
\textbf{b)} that the intermediate transformer representations are already quite powerful discriminators \cite{xin-etal-2020-deebert}.
Consequently, CascadeXML extracts features for classification at the coarser meta-label resolution from earlier transformer layers
and uses the last layer only for classification at the extreme resolution. The advantage of this is twofold: 
First, we postulate that the gain in classification accuracy due to enabling tree resolution-specific attention maps 
outweighs the loss in accuracy due to the slightly less powerful representations at the earlier transformer layers. Second,
this ensures that the representation capacity of the later layers is exclusively available to the more difficult tasks.

\begin{figure}[!t]
    \centering
    \begin{adjustbox}{width=\textwidth}
    \input{FigureCascade.tex}
    \end{adjustbox}
    \caption{Overview over CascadeXML. The meta-classifiers $W_1$ - $W_3$ use intermediate BERT features $\levelemb{t}$
    to discard all meta-labels except the highest scoring at that level (marked in red), so their descendants (gray)
    need not be considered at the next level. 
    More details in \autoref{sec:method} and \autoref{alg:training-step}.
    }
    \label{fig:diagram}
\end{figure}
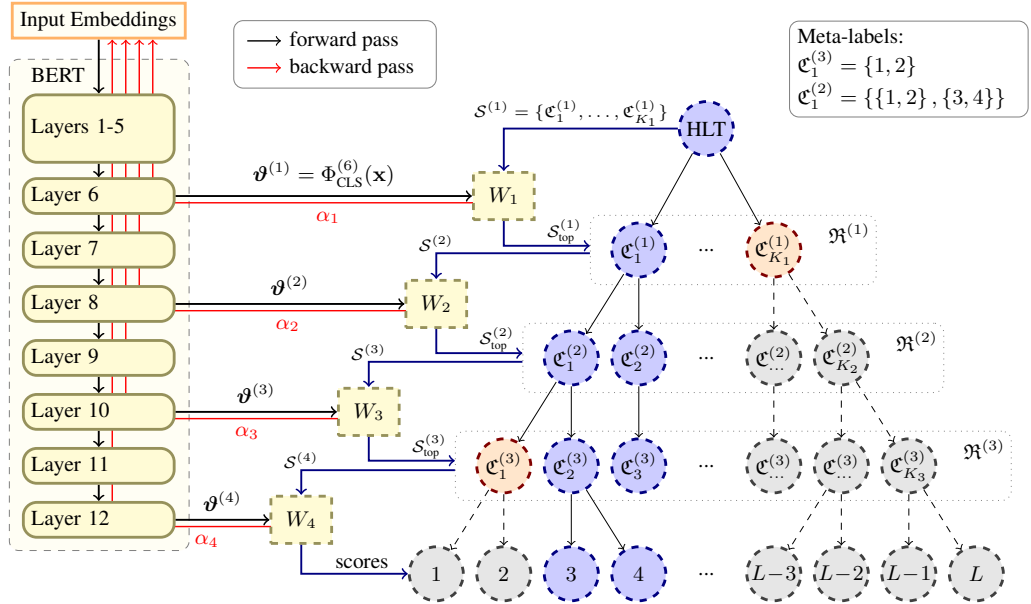

The main idea of CascadeXML is illustrated in \autoref{fig:diagram}. An input text is propagated through the transformer model until the first shortlisting level is reached. 
There, the embedding $\levelemb{1}$ corresponding to the [CLS] token is extracted, and used to determine scores for each level-1 meta-label. The highest-scoring meta-labels are selected to create a shortlist of candidate labels $\shortlist{1}_{\text{top}}$. 
The input is propagated further to perform another shortlisting step at the next resolution, at which a new [CLS] embedding  $\levelemb{2}$ is extracted. 
Crucially, $\levelemb{2}$ is the result of a separate self-attention operation, and its attention map is a differently weighted combination of the token embeddings as compared to $\levelemb{1}$. 
The new embedding $\levelemb{2}$ is used to refine the shortlist by selecting the top-scoring level-2 meta-labels that are children of the shortlisted level-1 meta-labels.
This process is repeated until at the final layer of the transformer, where classification is performed at the full label resolution.

\paragraph{Contributions}
In this paper, our contributions are the following:
\begin{itemize} 

\item We introduce a novel paradigm which can harness the multi-layered architecture of transformers for learning representations at multiple resolutions in an HLT. This is in contrast to the existing approaches which treat transformer as merely a black box encoder,
\item Implemented via CascadeXML, it is an end-to-end trainable multi-resolution learning pipeline which is quick to train and simple to implement. It eliminates the need of \textit{bells and whistles} like multi-stage procedures, reclustering, and bootstrap training to achieve state-of-the-art performance, and 
\item CascadeXML not only scales to dataset with millions of labels but also improves the current state-of-the-art in terms of prediction performance by upto $\sim 6\%$ on benchmark datasets. In addition to this, it is computationally most efficient to train and clocks multi-fold improvement in inference time as compared to existing transformer-based approaches. 
\end{itemize}

%% file: FigureCascade.tex
\pgfdeclarelayer{background}
\pgfdeclarelayer{foreground}
\pgfsetlayers{background,main,foreground}

\begin{tikzpicture}[
font={\small}, 
squarednode/.style={rectangle, draw=orange!60, fill=yellow!5, very thick, minimum size=5mm},
bertlayer/.style={rectangle, draw=yellow!50!black, fill=yellow!20, very thick, minimum width=2cm, rounded corners=2mm, align=left,text width=2cm},
linear/.style={rectangle, draw=yellow!50!black, fill=yellow!20, very thick, minimum width=0.9cm, minimum height=0.7cm, dashed},
metalabel/.style={circle, draw=blue!50!black, fill=blue!20, very thick, minimum width=0.8cm, minimum height=0.8cm, dashed, inner sep=0pt},
inactivelabel/.style={circle, draw=gray!50!black, fill=gray!20, very thick, minimum width=0.8cm, minimum height=0.8cm, dashed, inner sep=0pt},
remlabel/.style={circle, draw=red!50!black, fill=orange!20, very thick, minimum width=0.8cm, minimum height=0.8cm, dashed, inner sep=0pt},
shortlist/.style={font={\scriptsize},yshift=-1mm},
gradient/.style={red,->,semithick},
shortlistprop/.style={blue!50!black,->,thick,text=black},
resolutionlevel/.style={rounded corners, draw=black!50, dotted, minimum height=1cm, align=right, text depth=0.5cm},
node distance=0.8cm
]
    \node[squarednode]    (input)                                              {Input Embeddings};
    
    \node[text width=2cm] (bert)  [below of=input]                             {BERT};
    \node[bertlayer,minimum height=1cm]      (bert1)  [below of=bert]          {Layers 1-5};
    \node[bertlayer, node distance=1cm]      (bert2)  [below of=bert1]         {Layer 6};
    \node[bertlayer]      (bert3)  [below of=bert2]                            {Layer 7};
    \node[bertlayer]      (bert4)  [below of=bert3]                            {Layer 8};
    \node[bertlayer]      (bert5)  [below of=bert4]                            {Layer 9};
    \node[bertlayer]      (bert6)  [below of=bert5]                            {Layer 10};
    \node[bertlayer]      (bert7)  [below of=bert6]                            {Layer 11};
    \node[bertlayer]      (bert8)  [below of=bert7]                            {Layer 12};
    \node                 (bert9)  [below of=bert8]                            {};

    \path (bert2)+(6cm,0) node (short1) [linear] {$W_1$};
    \path (bert4)+(5cm,0) node (short2) [linear] {$W_2$};
    \path (bert6)+(4cm,0) node (short3) [linear] {$W_3$};
    \path (bert8)+(3cm,0) node (short4) [linear] {$W_4$};

    \draw[->, thick] (bert2.east) -- node[above] (){$\levelemb{1} = \encoder[CLS]^{(6)}(\instance)$} (short1.west); 
    \draw[->, thick] (bert4.east) -- node[above] (){$\levelemb{2}$} (short2.west); 
    \draw[->, thick] (bert6.east) -- node[above] (){$\levelemb{3}$} (short3.west); 
    \draw[->, thick] (bert8.east) -- node[above] (){$\levelemb{4}$} (short4.west); 

    \path (bert1)+(9cm,0) node (treeroot) [metalabel] {HLT};
    \path (bert3)+(8cm,0) node (r11) [metalabel] {$\cluster{1}[1]$};
    \path (bert3)+(9cm,0) node (r12) [] {...};
    \path (bert3)+(10cm,0) node (r12) [remlabel] {$\cluster{1}[\nummeta_1]$};

    \path (bert5)+(7cm,0) node (r21) [metalabel] {$\cluster{2}[1]$};
    \path (bert5)+(8cm,0) node (r22) [metalabel] {$\cluster{2}[2]$};
    \path (bert5)+(9cm,0) node (r23) [] {...};
    \path (bert5)+(10cm,0) node (r24) [inactivelabel] {$\cluster{2}[...]$};
    \path (bert5)+(11cm,0) node (r25) [inactivelabel] {$\cluster{2}[\nummeta_2]$};

    \path (bert7)+(6cm,0) node (r31) [remlabel] {$\cluster{3}[1]$};
    \path (bert7)+(7cm,0) node (r32) [metalabel] {$\cluster{3}[2]$};
    \path (bert7)+(8cm,0) node (r33) [metalabel] {$\cluster{3}[3]$};
    \path (bert7)+(9cm,0) node (r34) [] {...};
    \path (bert7)+(10cm,0) node (r35) [inactivelabel] {$\cluster{3}[...]$};
    \path (bert7)+(11cm,0) node (r36) [inactivelabel] {$\cluster{3}[...]$};
    \path (bert7)+(12cm,0) node (r37) [inactivelabel] {$\cluster{3}[\nummeta_3]$};

    \path (bert9)+(5cm,0) node (r41) [inactivelabel] {$1$};
    \path (bert9)+(6cm,0) node (r42) [inactivelabel] {$2$};
    \path (bert9)+(7cm,0) node (r43) [metalabel] {$3$};
    \path (bert9)+(8cm,0) node (r44) [metalabel] {$4$};
    \path (bert9)+(9cm,0) node (r45) [] {...};
    \path (bert9)+(10cm,0) node (r46) [inactivelabel] {$\numlabels\!-\!3$};
    \path (bert9)+(11cm,0) node (r47) [inactivelabel] {$\numlabels\!-\!2$};
    \path (bert9)+(12cm,0) node (r48) [inactivelabel] {$\numlabels\!-\!1$};
    \path (bert9)+(13cm,0) node (r49) [inactivelabel] {$\numlabels$};

    \draw[->] (treeroot) -- (r11);
    \draw[->] (treeroot) -- (r12);
    \draw[->] (r11) -- (r21);
    \draw[->] (r11) -- (r22);
    \draw[->, dashed] (r12) -- (r24);
    \draw[->, dashed] (r12) -- (r25);
    \draw[->] (r21) -- (r31);
    \draw[->] (r21) -- (r32);
    \draw[->] (r22) -- (r33);
    \draw[->, dashed] (r24) -- (r35);
    \draw[->, dashed] (r25) -- (r36);
    \draw[->, dashed] (r25) -- (r37);

    \draw[->, dashed] (r31) -- (r41);
    \draw[->, dashed] (r31) -- (r42);
    \draw[->] (r32) -- (r43);
    \draw[->] (r32) -- (r44);
    \draw[->, dashed] (r36) -- (r46);
    \draw[->, dashed] (r36) -- (r47);
    \draw[->, dashed] (r37) -- (r48);
    \draw[->, dashed] (r37) -- (r49);

    \begin{pgfonlayer}{background}
        \node (bt) at ([shift=(0:-.2)]bert.north west) {};
        \node (bte) at ([shift=(0:.2),yshift=-5]bert8.south east) {};
        \path[fill=yellow!5,rounded corners, draw=black!50, dashed]
            (bt) rectangle (bte);
        \draw[->, thick] (input) -- (bert1);
        \draw[->, thick] (bert1) -- (bert2);
        \draw[->, thick] (bert2) -- (bert3);
        \draw[->, thick] (bert3) -- (bert4);
        \draw[->, thick] (bert4) -- (bert5);
        \draw[->, thick] (bert5) -- (bert6);
        \draw[->, thick] (bert6) -- (bert7);
        \draw[->, thick] (bert7) -- (bert8);


        \draw[gradient] ($(short1)-(0,0.1)$) --node[below] (){$\alpha_1$} ($(bert2)+(0.8,-0.1)$) -- ($(input.south)+(0.8,0)$);
        \draw[gradient] ($(short2)-(0,0.1)$) --node[below] (){$\alpha_2$} ($(bert4)+(0.6,-0.1)$) -- ($(input.south)+(0.6,0)$);
        \draw[gradient] ($(short3)-(0,0.1)$) --node[below] (){$\alpha_3$} ($(bert6)+(0.4,-0.1)$) -- ($(input.south)+(0.4,0)$);
        \draw[gradient] ($(short4)-(0,0.1)$) --node[below] (){$\alpha_4$} ($(bert8)+(0.2,-0.1)$) -- ($(input.south)+(0.2,0)$);

        \node[resolutionlevel, text width=4.0cm] at ($(bert3)+(9.4cm,0mm)$) (resolution1) {$\resolution{1}$};
        \node[resolutionlevel, text width=6.0cm] at ($(bert5)+(9.4cm,0mm)$) (resolution2) {$\resolution{2}$};
        \node[resolutionlevel, text width=8.0cm] at ($(bert7)+(9.4cm,0mm)$) (resolution3) {$\resolution{3}$};
    \end{pgfonlayer}
    
    \path[rounded corners, draw=black!50] (2cm,0mm) rectangle (5cm,-1cm);
    \draw[->,semithick] (2.2cm,-3mm) -- (2.7cm,-3mm) node[right] {forward pass};
    \draw[->,semithick, red,text=black] (2.2cm,-7mm) -- (2.7cm,-7mm)node[right] {backward pass};

    \draw[shortlistprop] (treeroot) -|node[above, xshift=1cm,shortlist] (){$\shortlist{1} = \{\cluster{1}[1], \ldots, \cluster{1}[\nummeta_1]\}$} (short1);
    \draw[shortlistprop] (short1) |-node[above,xshift=0.9cm,shortlist] (){$\shortlist{1}_{\text{top}}$} ([yshift=1mm]resolution1);
    \draw[shortlistprop] ([yshift=-1mm]resolution1) -|node[above,shortlist] (){$\shortlist{2}$} (short2);
    \draw[shortlistprop] (short2) |-node[above,xshift=0.9cm,shortlist] (){$\shortlist{2}_{\text{top}}$} ([yshift=1mm]resolution2);
    \draw[shortlistprop] ([yshift=-1mm]resolution2) -|node[above,shortlist] (){$\shortlist{3}$} (short3);
    \draw[shortlistprop] (short3) |-node[above,xshift=0.9cm,shortlist] (){$\shortlist{3}_{\text{top}}$} ([yshift=1mm]resolution3);
    \draw[shortlistprop] ([yshift=-1mm]resolution3) -|node[above,shortlist] (){$\shortlist{4}$} (short4);
    \draw[shortlistprop] (short4) |-node[above,xshift=0.9cm,yshift=-0.5mm] (){scores} (r41);
    
    \node[text width=3.3cm,rounded corners, draw=black!50,anchor=north] (ml) at(12cm, 0cm) {Meta-labels:\\[2pt]$\cluster{3}_1 = \set{1, 2}$\\[1pt]$\cluster{2}_1 = \set{\set{1, 2}, \set{3, 4}}$};
\end{tikzpicture}

%% file: Background.tex
\section{Notation \& Preliminaries}

\textbf{Problem setup:} In extreme classification, the input text instance (document or query) $\instance \in \featurespace$ is mapped to a subset of labels $\labelset \subset \intrange[\numlabels]$ out of the $\numlabels$ available labels
identified through the integers $\intrange[\numlabels] \coloneqq \set{1, \ldots, \numlabels}$, such that $\numlabels
\sim 10^6$. Usually, we identify the labels through a binary vector $\labelvec \in
\set{0,1}^{\raisemath{-2pt}{\numlabels}}$, where $y_j = 1 \Leftrightarrow j \in \labelset$. 
The instances and labels are jointly distributed according to an unknown distribution $(\instance, \labelvec) \sim \mathds{P}$ of which we have a i.i.d. sample $\mathcal{D} = \left\{ \left(\instance^{i}, \labelvec^{i}\right): i \in [n] \right\}$ available as a training set. 
Even though the number of labels is large, the label vectors are sparse. 

As part of the model pipeline, a text vectorizer $\encoder: \featurespace \mapsto \mathbb{R}^d$ projects the documents to a $d$-dimensional embedding space. These vectorizers can be simple bag-of-words (BOW) or TF-IDF transformations $\encoder[tf-idf]$, or a learnable model $\encoder[dnn]$ that encodes the documents to the embedding space using weights learnt over the dataset. The final classification of an instance $\instance$ is achieved by computing per
label scores $ \langle \textbf{w}_l, \Phi(\instance)\rangle$ using label-wise weight vectors $\textbf{w}_l \in \reals^d$ which are
either (i) \emph{jointly} learnt with $\encoder$ in a deep architecture \cite{Chang20, Dahiya21, Jiang2021, Mittal21,
Ye20, you18}, or (ii) \emph{solely} learnt in shallow classifiers \cite{Babbar17, yu2020pecos}.

\textbf{Transformer Encoder: } More recent approaches in extreme classification have achieved state-of-the-art
performance using pre-trained transformers as text vectorizers \cite{Jiang2021, Ye20, zhang2021fast}. Popular
transformer encoders for NLP \cite{Bert, RoBERTa, XLNet} consist of an embedding layer, followed by a series of stacked
self-attention layers and task-specific heads that can be fine-tuned along with the encoder. For a sequence of $m$ input
words,
each layer $a$ in these models produces a sequence of embeddings
$\left(\encoder^{(a)}_0(\instance), \ldots, \encoder^{(a)}_m(\instance) \right)$, where $\encoder^{(a)}_i(\instance) \in
\reals^d$ is a representation for the $i$'th token in the input sequence at the $a$'th layer, and
$\encoder[CLS]^{(a)}(\instance) \coloneqq \encoder^{(a)}_0(\instance)$ is an embedding corresponding to a special [CLS]
token that has been pre-trained to capture a representation of the entire input \cite{Bert, RoBERTa}.

\textbf{Label Shortlisting:} Computing scores $\langle \textbf{w}_l, \Phi(\instance)\rangle$ for each label $l \in
[\numlabels]$ in an extremely large output space is computationally very demanding, especially when the feature space
$\reals^d \ni \Phi(\instance)$ is high dimensional to allow for expressive features. To alleviate this problem, the
labels $\numlabels$ are grouped together under $\nummeta$ \emph{meta-labels} $ \set{\clustersymb_1,
\ldots, \clustersymb_\nummeta} \eqqcolon \resolutionsymb $, with $\nummeta \approx \mathcal{O}(\sqrt{\numlabels})$. Each meta-label is represented
as the set of its child labels, $\clustersymb_k = \{l
\in [\numlabels]: \text{$l$ is child of $k$}\}$ and each target set $\labelset \subset [\numlabels]$ corresponds to a
meta-target $\tilde{\labelset} = \resolutionsymb(\labelvec) \coloneqq \set{\clustersymb_k \mid \exists i \in \mathsf{y}: i \in
\clustersymb_k}$. A meta-label $\clustersymb_l$ is relevant to a data point if it contains at least one relevant label.

Ideally, the labels grouped under one meta-label should be semantically similar to ensure the common representations learnt for meta- and extreme- tasks are relevant to both tasks \cite{Dahiya21}.
This has the added advantage that the negative labels selected by the shortlisting procedure will be the most confusing ones, leading to hard negative sampling and improved learning of the extreme classifier \cite{Reddi2018}. 

The label vectors are sparse to the extent that the much coarser meta-labels still contain mostly zeros, $|\tilde{\labelset}| \ll \nummeta$. 
This means that it is enough to explicitly calculate the scores only for the descendants of relevant meta-labels.
In practice, an algorithm will first predict scores for each meta-label and select the top-scoring ones as a \emph{shortlist} for further classification. 
The meta- and extreme-classification tasks can either be learnt sequentially \cite{Chang20, Dahiya21} or jointly through dynamic label-shortlisting \cite{Jiang2021, kharbanda2021embedding}.

For extremely large label spaces, 
a single label resolution for shortlisting is not optimal \cite{zhang2021fast}.
It is either too large that the meta-task itself becomes very expensive, or its individual clusters are so large that the shortlist contains too many candidate labels \cite{zhang2021fast}. 
Thus, methods like AttentionXML and XR-Transformer use a multi-level shortlisting procedure based on a \emph{Hierarchical Label Tree (HLT)}.

\textbf{Label Trees:} An HLT $\hlt = \set{\resolution{1}, \ldots, \resolution{\numlevels}}$ is a hierarchical clustering of labels into $\numlevels$ levels of successively refined clusters $\resolution{t} = \left\{\cluster{t}[1], \ldots, \cluster{t}[\nummeta_t] \right\}$. 
The clusters $\cluster{t}[k]$ of each level form a disjoint partitioning of the clusters of the next, $\cluster{t}[k] \subset \resolution{t+1}$, with the convention that the $\numlevels+1^{\text{th}}$ level is the full label space $\resolution{\numlevels+1} = [\numlabels]$. 
As such, each cluster $\cluster{\numlevels}[k]$ (or meta-label) at level $\numlevels$ is represented by a set of labels, each cluster at level $\numlevels-1$ is a set of level-$\numlevels$ meta-labels, and thus a set of sets of labels, and so on. In practice, label representations $\mathbf{z}_l$ as an aggregate of their positive instance features are leveraged to create an HLT. 
Typically, $\mathbf{z}_l$ is constructed using $\encoder[tf-idf]$ as
\begin{equation*}
    \mathbf{z}_{l} = \frac{\mathbf{v}_l}{\left\| \mathbf{v}_l \right\|},\ \ \textrm{where} \ \mathbf{v}_l = \sum_{\mathclap{i:\labelvec^{(i)}_{l}=1}} \encoder[tf-idf](\instance^{(i)}).
\end{equation*} 
The HLT is often constructed by recursively partitioning the label-space using balanced k-means clustering in a top-down fashion \cite{Chang20, Jiang2021, you18, zhang2021fast}. For example, an HLT constructed using balanced 2-means clustering has $\lfloor \log_2 \numlabels \rfloor$ levels, out of which we select $\numlevels$ levels corresponding to different meta-label resolutions. 

%% file: Method.tex
\section{Method: CascadeXML}
\label{sec:method}

At its core, the CascadeXML model consists of three components: A pre-trained language model $\encoder$ which allows
to extract a document representation $\encoder[CLS]^{(a)}(\instance) \in \reals^d$ at different layers $a \in [A]$, an HLT
$\hlt = \set{\resolution{1}, \ldots, \resolution{\numlevels}}$ that divides the label space into $\nummeta_1 < \ldots <
\nummeta_\numlevels < \numlabels=\nummeta_{\numlevels+1}$ successively refined clusters, and a set of linear classifiers
$\set{\mathbf{W}^{(t)} = [\mathbf{w}_1^{(t)}, \ldots, \mathbf{w}_{\nummeta_t}^{(t)}]\in \reals^{d \times \nummeta_t}: t
\in [\numlevels+1]}$. 
For each meta-classifier, we select a layer in order to extract resolution-specific feature embeddings\linebreak $\levelemb{t} \coloneqq \encoder[CLS]^{(a_t)}(\instance)$ with $1 < a_1 < \ldots < a_\numlevels < A = a_{\numlevels+1}$.

\textbf{Refining and Coarsening Label Vectors:} 
In order to transfer the representation of a shortlist of a label vector between different levels of the HLT, we need to define two operations. 
The first is refining the resolution of a shortlist: Given a shortlist $\shortlistsymb = \{\cluster{t}[k_1], \ldots, \cluster{t}[k_s]\} \subset \resolution{t}$ of clusters of the resolution at level $t$, we want to calculate the representation of this shortlist in the next level, denoted $\refinecluster{t}(\shortlistsymb)$. 
Since each cluster is defined as the set of its descendants, this is achieved by taking the union over the clusters in the list. 
The second operation is to find the cover of meta-labels that envelope a given set of labels, $\coarsencluster{t}(\shortlistsymb)$.
This is achieved by identifying each of the clusters in the level above for which there exists at least one descendant in the shortlist. This leads to
\begin{equation}
    \refinecluster{t}(\shortlistsymb) = \bigcup_{C \in
\shortlistsymb} C \subset \resolution{t+1}, \qquad
\coarsencluster{t}(\shortlistsymb) = 
\left\{\cluster{t-1}[k] \in \resolution{t-1} \mid \shortlistsymb \cap  \cluster{t-1}[k] \ne \emptyset \right\}.
\end{equation} 
We define $\labelsatlevel{t}{\labelvec} = \left(
\coarsencluster{t+1} \circ \ldots \circ \coarsencluster{T} \right)(\mathsf{y})$ to be the level-$t$ cover of the label set $\labelset \subset [\numlabels]$.

\textbf{Multi-resolution Dynamic Label Shortlisting:}
We adapt dynamic label shortlisting \cite{Jiang2021} for multi-resolution training approaches.
Formally, given a shortlist $\shortlist{t}$ of $s$ clusters $\{\cluster{t}[k_1], \ldots, \cluster{t}[k_s]\} \eqqcolon \shortlist{t} \subset \resolution{t}$ at level $t$, and the corresponding classification features $\levelemb{t}$, a new shortlist $\shortlist{t+1} \subset \resolution{t+1}$ is generated as follows:
The $k$ highest-scoring meta-labels are selected into $\shortlist{t}_{\text{top}}$ and the next shortlist $\shortlist{t+1}$ is then given as the refinement:
\begin{equation}
    \shortlist{t}_{\text{top}} = \set{\cluster{t}[k_j]: j \in \operatorname{Top}_k\!\left(\left\langle \mathbf{w}^{(t)}_{k_i}, \levelemb{t} \right\rangle \right)}, \quad \shortlist{t+1} = \refinecluster{t}\!\left(\shortlist{t}_{\text{top}}\right)
\end{equation}
At the first level, however, the shortlist is initialized to consider all meta labels $\shortlist{1} = \resolution{1}$. 
It is of essence that all true positive (meta-)labels appear at every level during training time \cite{Chang20, Jiang2021, you18}.
However, this is difficult to ensure implicitly due to imperfect recall rate of classifiers.
Hence, during training all meta-labels corresponding to ground-truth labels are added to the candidate set, leading to
\begin{equation}
    \shortlist{t+1}_{\text{train}} =  \refinecluster{t}(\shortlist{t}_{\text{top}}) \cup \labelsatlevel{t+1}{\labelvec}.
    \label{teacher-forcing}
\end{equation}
\input{StepAlgo}
Consequently, a single training step in CascadeXML looks as given in \autoref{alg:training-step}.

\textbf{Training Objective:} 
At each level of the tree, the goal is to correctly identify the most likely meta-labels, which can be achieved through minimization of a one-vs-all loss, such as BCE
\begin{equation}
\abovedisplayskip=2pt
    \mathcal{L}^{(t)}(\instance, \labelvec) = \frac{1}{|\shortlist{t}|} \sum_{l \in \shortlist{t}} \mathcal{L}^{(t)} \left(\langle \mathbf{w}^{(t)}_{k_l}, \levelemb{t}(\instance) \rangle, \labelsatlevel{t}{\labelvec}_l\right).
\end{equation}
The overall objective is a weighted sum of the individual layer losses
\begin{equation}
\abovedisplayskip=2pt
    \mathcal{L}(\instance, \labelvec) = \sum^{\numlevels + 1}_{t=1}\alpha^{(t)} \mathcal{L}^{(t)}(\instance, \labelvec) \, ,
\end{equation}
where level $\numlevels+1$ corresponds to the extreme classification task.
To balance out different loss scales as a result of normalization by varied shortlist sizes, we re-scale losses from multiple resolutions using a simple factor $\alpha^{(t)} = |\shortlist{t}|/\min_{t \in [T+1]}(|\shortlist{t}|)$.

\textbf{Inference: } The algorithm for inference remains the same as training, except we do not teacher-force true (meta-)labels $\resolution{t}(\mathsf{y})$ during inference time (Eqn. \ref{teacher-forcing}). 
The inference consists of one forward pass through the transformer backbone, followed by successive shortlisting at all $T$ resolutions, and final extreme classification. For a $b$-way branching tree with $\log_b \numlabels$ levels, each level $t$ has $b^t$ meta-labels, of which $|\shortlist{t}| \propto \sqrt{b^t}$ are selected for the shortlist. Therefore, the time complexity is
\begin{equation*}
    \mathcal{O}(\mathcal{T}_{\text{dnn}} + (|\shortlist{1}| + |\shortlist{2}| + ... + |\shortlist{\numlevels+1})|d) = 
    \mathcal{O}(\mathcal{T}_{\text{dnn}} + (\sum_{t=1}^{\log_b \numlabels} \sqrt{b^t})d) = \mathcal{O}(\mathcal{T}_{\text{dnn}} + d \sqrt{\numlabels})
\end{equation*}
where $\mathcal{T}_{\text{dnn}}$ denotes the time taken to compute transformer embeddings for an input.

%% file: StepAlgo.tex

\begin{algorithm}
\SetKwInput{KwData}{Input}
\SetKwInput{KwResult}{Return}
\DontPrintSemicolon
\SetNoFillComment
\caption{Training step in \textsc{CascadeXML}}\label{alg:Cascade}
\label{alg:training-step} 
\KwData{instance $\instance$, labels $\labelvec$, clusters $\clustersymb$, features $\encoder$, classifiers $\mathbf{W}$,
weights $\boldsymbol{\alpha}$}
$\levelemb{t} \gets \encoder[CLS]^{(t)}(\instance) \; \forall t \in [\numlevels+1]$ \hfill \tcc{Forward pass through Transformer}
$\textrm{loss} \gets 0$\;
$\shortlist{1} \gets \cluster{1}$  \hfill \tcc{Initialize shortlist}
\For{$t \; \textrm{in} \; 1,2,...,\numlevels$}{
    $\labelvec^{(t)} \gets \resolution{t}(\labelvec)$ \hfill \tcc{Ground-truth meta-labels}
    $p_i \gets \langle \mathbf{w}^{(t)}_{k_i}, \levelemb{t} \rangle \; \forall i \in \shortlist{t}$\ \hfill \tcc{Sparse prediction}
    $\text{loss} \gets \text{loss} + \frac{\alpha_t}{|\shortlist{t}|} \sum_{l \in \shortlist{t}} \mathcal{L}^{(t)} \left(p_l, y^{(t)}_l\right)$

    $\shortlist{t}_{\text{top}} \gets \set{\cluster{t}[k_j]: j \in \operatorname{Top}_k\left(\set{p_i \mid i \in \shortlist{t}} \right)}.$ \hfill \tcc{Top-K selection}
        $\shortlist{t+1} \gets \refinecluster{t}(\shortlist{t}_{\text{top}}) \cup \resolution{t+1}(\mathbf{y})$ \hfill \tcc{Add ground-truth positives}
}
$p_i \gets \langle \mathbf{w}^{(\numlevels+1)}_{k_i}, \levelemb{\numlevels+1} \rangle \quad \forall i \in \shortlist{\numlevels+1}$\ \hfill \tcc{Final prediction}
$\text{loss} \gets \text{loss} +\frac{\alpha_{T+1}}{|\shortlist{T+1}|} \sum_{l \in \shortlistsymb} \mathcal{L}^{(T+1)} \left(p_l, y_l\right)$ \;
adjust $\encoder$ and $\mathbf{W}$ to reduce loss
\end{algorithm}

%% file: Experiments.tex
\section{Main results}
\paragraph{Benchmark datasets \& Evaluation Metrics} 
We benchmark the performance of CascadeXML on standard datasets and evaluation metrics, as shown in \autoref{tbl:results} and \autoref{tbl:NE_results}.
In all, we use 5 public XMC datasets \cite{ExtremeRepository, mcauley2013hidden}, the statistics of which are specified in appendix A.1. 
For all our experiments we use the same raw text input, train-test split and sparse feature representations as used in \cite{you18, Jiang2021, zhang2021fast}. 
We use Precision@k (P@k) and Propensity-scored P@k (PSP@k) \cite{jain2016extreme}, which focuses more on the model's performance on tail labels, as the evaluation metrics. 

\paragraph{Baseline Methods} We compare CascadeXML with strong baselines:  DiSMEC \cite{Babbar17}, Parabel \cite{Parabel}, eXtremeText \cite{Wydmuch18}, Bonsai \cite{Khandagale19}, XR-Linear \cite{yu2020pecos}, AttentionXML, X-Transformer \cite{Chang20}, LightXML and XR-Transformer. 
The baseline results for these methods were obtained from \cite[Table: 2]{zhang2021fast}. 
It may be noted that, in order to incorporate the global statistical information of the dataset,  XR-Transformer concatenates sparse \textit{tf-idf} features to those learnt via deep networks (denoted $\encoder[dnn]$) as:
\begin{equation*}
   \encoder[cat] (\instance) = \left[\frac{\encoder[dnn](\instance)}{\|\encoder[dnn](\instance)\|} , \frac{\encoder[tf-idf](\instance)}{\|\encoder[tf-idf](\instance)\|} \right]
\end{equation*}
For fair evaluation, we present separate comparison with methods which use (i) features from a DNN encoder only, and (ii) those obtained by combining DNN and tf-idf features.

\paragraph{Model Ensemble Settings}
For CascadeXML, we follow the ensemble settings of LightXML and XR-Transformer. Specifically, we ensemble a model each using BERT \cite{Bert}, RoBerta. \cite{RoBERTa} and XLNet \cite{XLNet} for Wiki10-31K and AmazonCat-13K. For the larger datasets, we make an ensemble of three BERT models with different seeds for random initialization. On the other hand, AttentionXML uses ensemble of 3 models and X-Transformer uses 9 model ensemble with BERT, RoBerta, XLNet large models with three difference indexers. 

\input{TableResults.tex}
\textbf{Empirical Performance:} 
As demonstrated in \autoref{tbl:results}, CascadeXML, which leverages the multi-layered transformer architecture to learn label resolution specific feature representations, is empirically superior to all existing XMC approaches whether using: (i) tf-idf features, or (ii) DNN-based dense features, or (iii) both. 
This is despite CascadeXML being a simple end-to-end trainable algorithm without using any bells or whistles like multi-stage procedures \cite{you18, zhang2021fast}, reclustering \cite{zhang2021fast} or bootstrapped training \cite{you18, zhang2021fast}.
For example, on Amazon-3M dataset, CascadeXML significantly outperforms all DNN-based approaches while taking only 24 hours to train on a single Nvidia A100 GPU. In contrast, X-Transformer needs to ensemble 9 models and 23 days to train on 8 GPUs to reach the empirical performance (c.f. page 2, \cite{zhang2021fast}), and LightXML remains unscalable. 

\input{TableNEResults.tex}
We report results on Wiki10-31K and AmazonCat-13K in \autoref{tbl:NE_results}. 
As a standard practice in the domain \cite{Chang2020Pre-training, Jiang2021, zhang2021fast}, we use 256 tokens as an input sequence length to the model after truncation.
We find this sequence length to be sufficient for competent empirical performance and hence, we do not witness any empirical benefits of leverage sparse tf-idf features for these datasets.
Notably, CascadeXML achieves best results on AmazonCat-13K dataset and performs at par with XR-Transformer without leveraging sparse tf-idf features. Thus, CascadeXML's unique multi-resolution approach is a favorable choice not only for datasets with extremely large output spaces, but also for datasets with tens of thousand labels.\\

On the concatenated feature space, CascadeXML also significantly outperforms XR-Transformer on 8 out of 9 dataset-metric combination (ref: \autoref{tbl:results}) with notable $\sim 6\%$ improvement on Wiki-500K dataset for P@5 metric. 
As XR-Tranformer learns transformer features for label resolutions only up to $2^{15}$ clusters \cite[Table 6]{zhang2021fast}, incorporating tf-idf features via a computation-intensive XR-Linear \cite{yu2020pecos} for final extreme classification task becomes an integral part of their pipeline.
On the other hand, CascadeXML does not have any such inherent limitation, and even without tf-idf features, it outperforms XR-Transformer on 7 out of 9 dataset-metric combinations (ref: \autoref{tbl:results}).

\section{Computational cost \& Ablation results}

\paragraph{Single Model Comparison \& Training Time}
Empirical results and training time of a single CascadeXML model has been compared to single instance performance of state-of-the-art DNN based XMC approaches in \autoref{tbl:single_model}.
Notably, CascadeXML, without leveraging any sparse tf-idf features, performs at par with XR-Transformer which benefits from using the extra statistical information about the entire (not truncated) data point. 
When comparing training time, CascadeXML is up to an order of magnitude faster than LightXML on a single GPU. 
In multi-GPU setting, CascadeXML trains in 7.2 hours on Wiki-500K dataset using only 4 NVidia V100 GPUs as compared to XR-Transformer and AttentionXML which require 12.5 hours on 8 GPUs. 
CascadeXML not only reduces the training time across datasets but also requires half the number of GPUs to do so. 
This proves that CascadeXML is significantly more compute efficient as compared to previous DNN based XMC approaches. 

\input{TableSingleResults.tex}

\paragraph{Inference Time}
CascadeXML clocks the fastest inference speed as compared to all previous DNN-based XMC models, as shown in \autoref{tab:inference_speed}. Notably, CascadeXML is $\sim1.5 \times$ and $\sim2\times$ faster at inference time compared to LightXML and XR-Transformer respectively.
\input{TableInferenceTime.tex}

\paragraph{Performance on Tail Labels}
Performance of CascadeXML on tail labels has been compared to baseline XMC methods in \autoref{tbl:psp_results}. 
We note that for smaller datasets - Wiki10-31K and AmazonCat-13K - PfastreXML \cite{jain2016extreme} significantly outperforms all other methods on PSP metrics. This is to be expected, as this method specifically optimizes for tail labels.
However, for larger datasets - Amazon-670K and Wiki-500K - CascadeXML outperforms the strong baselines like XR-Transformer and AttentionXML by 5-8\% on PSP@3 and PSP@5 metrics while also significantly outperforming PfastreXML over these datasets. 
These results conclude that our unique end-to-end multi-resolution training approach is not only empirically superior to previous approaches in P@K metrics but also outperforms strong XMC baselines in performance over tail labels.

\input{TablePSPResults.tex}

\input{TableAblations.tex}
\paragraph{Impact of label clusters size} Earlier works have argued that using fine-grained clusters leads to better model performance \cite{Jiang2021, kharbanda2021embedding, Mittal21}. 
With increased number of label clusters at the penultimate level of HLT, the multi-resolution tasks tend to get more in-sync with each other \cite{kharbanda2021embedding} and hence enables transfer learning across different stages of learning pipelines \cite{Dahiya21, Mittal21}.
Our mutli-resolution training enables CascadeXML to use $2^{16}$ label clusters for penultimate label resolution,  double that of XR-Transformer. As \autoref{tab:rescale_loss} shows, increasing the resolution at the last shortlisting step results in non-trivial improvements, highlighting the need for architectures that can efficiently handle high-resolution meta-labels.

\paragraph{Impact of re-scaling per-resolution loss} 
For extreme datasets, we witness more efficient training by having a larger shortlist size for finer label resolutions (see appendix A.2.2). 
This results in varied loss scales across label resolutions leading to one resolution dominating the training. 
Comparing the ``Default'' and ``w/o Rescaling'' rows for respective datasets in \autoref{tab:rescale_loss}, indicates that rescaling as introduced in section 3 helps CascadeXML train more effectively.

%% file: TableResults.tex
\begin{table}[!h]
\begin{adjustbox}{width=\textwidth,center}
\begin{tabular}{c|ccc|ccc|ccc}
\toprule
Metrics  & P@1            & P@3            & P@5            & P@1            & P@3            & P@5      
& P@1            & P@3            & P@5    \\

\specialrule{0.70pt}{0.4ex}{0.65ex}
                {Datasets} & \multicolumn{3}{c}{\textbf{Wiki-500K}}                    & \multicolumn{3}{c}{\textbf{Amazon-670K}}                  & \multicolumn{3}{c}{\textbf{Amazon-3M}}                    \\
\specialrule{0.70pt}{0.4ex}{0.65ex}
DiSMEC          & 70.21          & 50.57          & 39.68          & 44.78          & 39.72          & 36.17          & 47.34          & 44.96          & 42.80          \\
Parabel$^\ast$         & 68.70          & 49.57          & 38.64          & 44.91          & 39.77          & 35.98          & 47.42          & 44.66          & 42.55          \\
eXtremeText     & 65.17          & 46.32          & 36.15          & 42.54          & 37.93          & 34.63          & 42.20          & 39.28          & 37.24          \\
Bonsai$^\ast$          & 69.26          & 49.80          & 38.83          & 45.58          & 40.39          & 36.60          & 48.45          & 45.65          & 43.49          \\
XR-Linear       & 65.59          & 46.72          & 36.46          & 43.38          & 38.40          & 34.77          & 47.40          & 44.15          & 41.87          \\
\specialrule{0.70pt}{0.4ex}{0.65ex}
 & \multicolumn{3}{c}{} & \multicolumn{3}{c}{DNN Features} & \multicolumn{3}{c}{} \\
\specialrule{0.70pt}{0.4ex}{0.65ex}
AttentionXML$^\ast$   & 76.74          & 58.18          & 45.95          & 47.68          & 42.70          & 38.99          & 50.86          & 48.00          & 45.82          \\
LightXML$^\ast$       & 77.89          & 58.98          & 45.71          & 49.32          & 44.17          & 40.25          & -              & -              & -              \\
CascadeXML$^\ast$ & \textbf{78.39} & \textbf{59.86} & \textbf{46.49} & \textbf{50.41} & \textbf{45.32} & \textbf{41.55} & \textbf{52.75} & \textbf{50.83} & \textbf{48.90}\\
\specialrule{0.70pt}{0.4ex}{0.65ex}
 & \multicolumn{3}{c}{} & \multicolumn{3}{c}{[DNN $\bigoplus$ tf-idf] Features} & \multicolumn{3}{c}{} \\
\specialrule{0.70pt}{0.4ex}{0.65ex}
X-Transformer$^{\ast\ast}$  & 77.09          & 57.51 & 45.28 & 48.07 & 42.96 & 39.12 & 51.20 & 47.81 & 45.07\\
XR-Transformer$^\ast$ & 79.40 & 59.02 & 46.25 & 50.11 & 44.56 & 40.64 & \textbf{54.20} & 50.81 & 48.26\\ 
CascadeXML$^\ast$ & \textbf{81.13} & \textbf{62.43} & \textbf{49.12} & \textbf{52.15} & \textbf{46.54} & \textbf{42.44} & 53.91 & \textbf{51.24} & \textbf{49.52} \\
\bottomrule
\end{tabular}
\end{adjustbox}
\vspace{1pt}

\caption{Comparison of CascadeXML to state-of-the-art methods on large scale benchmark datasets. 
`\textbf{*}' and `\textbf{**}' imply ensemble results of 3 models and 9 models respectively, and
`-' implies that the model does not scale for that dataset due to GPU memory constraints.}
\vspace{-1em}
\label{tbl:results}
\end{table}

%% file: TableNEResults.tex
\begin{wraptable}{r}{8.5cm}
\vspace{-1em}
\begin{adjustbox}{width=0.6\textwidth,center}
\begin{tabular}{c|ccc|ccc}
\toprule
Metrics  & P@1 & P@3 & P@5 & P@1 & P@3 & P@5 \\
\specialrule{0.70pt}{0.4ex}{0.65ex}
Datasets &  \multicolumn{3}{c}{\textbf{Wiki10-31K}} & \multicolumn{3}{c}{\textbf{AmazonCat-13K}} \\
\specialrule{0.70pt}{0.4ex}{0.65ex}
DiSMEC  & 84.13          & 74.72          & 65.94          & 93.81          & 79.08          & 64.06          \\
Parabel & 84.19          & 72.46          & 63.37          & 93.02          & 79.14          & 64.51          \\
eXtremeText & 83.66          & 73.28          & 64.51          & 92.50          & 78.12          & 63.51          \\
Bonsai   & 84.52          & 73.76          & 64.69          & 92.98          & 79.13          & 64.46          \\
XR-Linear & 85.75          & 75.79          & 66.69          & 94.64          & 79.98          & 64.79          \\
\specialrule{0.70pt}{0.4ex}{0.65ex}
 &  \multicolumn{6}{c}{DNN Features} \\
\specialrule{0.70pt}{0.4ex}{0.65ex}
AttentionXML$^\ast$ & 87.34          & 78.18          & 69.07          & 95.84          & 82.39          & 67.32          \\
LightXML$^\ast$ & \underline{89.67} & 79.06 & 69.87 & 96.77 & 83.98 & 68.63 \\
CascadeXML$^\ast$ & \textbf{89.74} & \underline{80.13} & 70.75 & \textbf{96.90} & \textbf{84.13} & \textbf{68.78}\\
\specialrule{0.70pt}{0.4ex}{0.65ex}
 & \multicolumn{6}{c}{[DNN $\bigoplus$ tf-idf] Features} \\
\specialrule{0.70pt}{0.4ex}{0.65ex}
X-Transformer$^{\ast\ast}$ & 88.26 & 78.51 & 69.68 & 96.48 & 83.41 & 68.19\\
XR-Transformer$^\ast$ & 88.69 & \textbf{80.17} & \underline{70.91} & \underline{96.79} & 83.66 & 68.04\\
CascadeXML$^\ast$ & 89.18 & 79.71 & \textbf{71.19} & 96.71 & \underline{84.07} & \underline{68.69} \\
\specialrule{0.70pt}{0.4ex}{0.65ex}
\end{tabular}
\end{adjustbox}
\caption{Comparison of CascadeXML to state-of-the-art methods on legacy datasets.
}
\vspace{-1em}
\label{tbl:NE_results}
\end{wraptable}

%% file: TableSingleResults.tex
\begin{table}[!h]
\begin{adjustbox}{width=0.8\textwidth,center}
\begin{tabular}{c|c|ccc|cc}
\toprule
\textbf{Dataset}    & \textbf{Method}                      & \textbf{P@1}   & \textbf{P@3}   & \textbf{P@5}   & $T^1_{\text{train}}$  & $T^{m}_{\text{train}}$\\ \midrule
& AttentionXML & 87.1  & 77.8  & 68.8  &   -   & 0.5  \\
Wiki10-31K & LightXML & 87.8  & 77.3  & 68.0  & 6.7  &  - \\
 & XR-Transformer (+ tf-idf) & 88.0  & 78.7  & 69.1  & 1.3  & 0.5  \\
& CascadeXML & 88.4  & 78.3  & 68.9  & \textbf{0.3}  & \textbf{0.1}\\ 
\midrule
 & AttentionXML & 75.1  & 56.5  & 44.4  & -    & 12.5 \\
Wiki-500K & LightXML& 76.3  & 57.3  & 44.2  & 89.6 &  - \\
& XR-Transformer (+ tf-idf) & 78.1  & 57.6  & 45.0  & 29.2 & 12.5 \\
& CascadeXML & 77.0 & 58.3  & 45.1 & \textbf{22.0} & \textbf{7.2} \\
\midrule
& AttentionXML & 45.7  & 40.7  & 36.9  &   -   & 8.1  \\
Amazon-670K & LightXML & 47.3  & 42.2  & 38.5  & 53.0 & - \\
 & XR-Transformer (+ tf-idf) & 49.1  & 43.8  & 40.0  & 8.1  & 3.4  \\
 & CascadeXML & 48.8 & 43.8  & 40.1  & \textbf{7.5}  & \textbf{3.0} \\ \bottomrule
\end{tabular}
\end{adjustbox}
\vspace{1.5pt}
\caption{Single model comparison of DNN based XMC approaches. $T^1_{train}$ denotes the training time as reported on a single GPU (Nvidia V100).
$T^{m}_{train}$ denotes multi-GPU training time on 8 GPUs for AttentionXML and XR-Transformer, and on only 4 GPUs for CascadeXML.}
\label{tbl:single_model}
\vspace{-2em}
\end{table}


%% file: TableInferenceTime.tex
\begin{table}[h]
\begin{adjustbox}{width=\textwidth,center}
\begin{tabular}{c|c|c|c|c|c}
\toprule
Dataset       & AttentionXML & X-Transformer & XR-Transformer & LightXML & CascadeXML\\
\midrule
Wiki10-31K & 20.0 & 48.1 & 39.1 & 27.1  & \textbf{12.8} \\
AmazonCat-13K & 14.4 & 47.6 & 26.1 & 24.1 &  \textbf{13.1} \\
Wiki-500K & 80.1$^\ast$ & 48.1 & 33.9 & 27.3 &  \textbf{16.0}\\
Amazon-670K & 76.0$^\ast$ & 48.0 & 30.9 & 23.3 &  \textbf{16.6}  \\
Amazon-3M & 130.5$^\ast$ & 50.2 & 35.2 & -  & \textbf{16.9}\\
\bottomrule
\end{tabular}
\end{adjustbox}
\vspace{1pt}
\caption{Comparison of CascadeXML w.r.t. inference time (in milliseconds per sample) with SOTA XMC methods. Inference times were recorded on a single Nvidia V100 GPU and a single CPU with a batch size of 1. Superscript \textbf{*} implies that model parallel was used with 8 GPUs for inference.}
\vspace{-2em}
\label{tab:inference_speed}
\end{table}

%% file: TablePSPResults.tex
\begin{table}[h]
\begin{adjustbox}{width=\textwidth,center}
\begin{tabular}{c|ccc|ccc|ccc|ccc}
\toprule
Methods  & PSP@1 & PSP@3 & PSP@5 & PSP@1 & PSP@3 & PSP@5 & PSP@1 & PSP@3 & PSP@5 & PSP@1 & PSP@3 & PSP@5 \\
\specialrule{0.70pt}{0.4ex}{0.65ex}
& \multicolumn{3}{c}{\textbf{Wiki10-31K}} & \multicolumn{3}{c}{\textbf{AmazonCat-13K}} & \multicolumn{3}{c}{\textbf{Amazon-670K}} & \multicolumn{3}{c}{\textbf{Wiki-500K}} \\
\midrule

DiSMEC           & 10.60 & 12.37 & 13.61 & 51.41 & 61.02 & 65.86 & 26.26 & 30.14 & 33.89 & 27.42 & 32.95 & 36.95 \\
ProXML           & 17.17 & 16.07 & 16.38 & 61.92 & 66.93 & 68.36 & 30.31 & 32.31 & 34.43 & - & - & - \\
PfastreXML       & \textbf{19.02} & \textbf{18.34} & \textbf{18.43} & \textbf{69.52} & \textbf{73.22} & 75.48 & 29.30 & 30.80 & 32.43 & 32.02 & 29.75 & 30.19 \\
Parabel          & 11.69 & 12.47 & 13.14 & 50.92 & 64.00 & 72.10 & 26.36 & 29.95 & 33.17 & 26.88 & 31.96 & 35.26 \\
Bonsai           & 11.85 & 13.44 & 14.75 & 51.30 & 64.60 & 72.48 & 27.08 & 30.79 & 34.11 & 27.46 & 32.25 & 35.48 \\
\midrule
& \multicolumn{12}{c}{DNN Features (Single Model)}\\
\midrule
XML-CNN & 9.39  & 10.00 & 10.20 & 52.42 & 62.83 & 67.10 & 17.43 & 21.66 & 24.42 & - & - & - \\
AttentionXML & 16.20 & 17.05 & 17.93 & 53.52 & 68.73 & 76.26 & 29.30 & 32.36 & 35.12 & 30.05 & 37.31 & 41.74 \\
XR-Transformer & 12.16	& 14.86 &	16.40 & 50.51 & 64.92 & 74.63 & 29.21 & 33.49 &	37.65 & 32.10 & 39.41 & 43.75\\
CascadeXML & 13.22 & 14.70 & 16.10 & 52.08 & 67.46 & 76.19 & 30.45 & 35.02 & 38.97 & 31.25 & 39.35 & 43.29 \\
\midrule
& \multicolumn{12}{c}{DNN Features (Ensemble Model)}\\
\midrule
AttentionXML$^\ast$ & 15.57 & 16.80 & 17.82 & 53.76 & 68.72 & 76.38 & 30.29 & 33.85 & 37.13 & 30.85 & 39.23 & 44.34 \\
CascadeXML$^\ast$ & 13.36 & 15.06 & 16.56 & 52.68 & 68.50 & 77.52 & \textbf{31.40} & \textbf{36.22} & 40.28 & 32.60 & 42.03 & 46.66\\
\midrule
& \multicolumn{12}{c}{[DNN $\bigoplus$ tf-idf] Features (Ensemble Model)}\\
\midrule
XR-Transformer$^\ast$ & 12.25	& 15.00 & 16.75 & 50.72 & 65.66 & 75.95 & 29.77 & 34.05 & 38.29 & \textbf{32.70} & 40.44 & 45.02\\
CascadeXML$^\ast$ & 13.32 & 15.35 & 17.45 & 51.39 & 66.81 & \textbf{77.58} & 30.77 & 35.78 & \textbf{40.52} & 32.12 & \textbf{43.15} & \textbf{49.37}\\

\bottomrule
\end{tabular}
\end{adjustbox}
\vspace{3pt}
\caption{Comparison of performance of state-of-the-art methods on tail labels on benchmark datasets.}
\vspace{-1em}
\label{tbl:psp_results}
\end{table}

%% file: TableAblations.tex
\begin{wraptable}{r}{8.0cm}
\begin{adjustbox}{width=0.55\columnwidth,center}

\begin{tabular}{c|c|ccc}
\toprule
Dataset & Model Setting & P@1 & P@3 & P@5 \\
\midrule 
& Default & \textbf{48.5} & \textbf{43.7} & \textbf{40.0}\\
Amazon-670K  & $2^{15}$ Clusters & 47.6 & 42.9 & 39.3\\
& w/o Rescaling  & 47.8 & 42.8 & 39.2 \\
\midrule
& Default & \textbf{76.9} & \textbf{58.4} & \textbf{45.2}\\
Wiki-500K  & $2^{15}$ Clusters & 76.6 & 58.1 & 44.9\\
& w/o Rescaling & 76.5 & 57.9 & 44.7\\
\midrule
 & Default & \textbf{51.3} & \textbf{49.0} & \textbf{46.9}\\
Amazon-3M  & $2^{15}$ Clusters & 50.8 & 48.7 & 46.5\\
& w/o Rescaling &  50.9 & 48.6 & 46.6\\

\bottomrule
\end{tabular}


\end{adjustbox}

\vspace{1pt}
\caption{Ablation experiments showing performance of a \textit{single} CascadeXML model with different model settings. Default implies $2^{16}$ clusters in penultimate label resolution and per-resolution loss re-scaling.}
\label{tab:rescale_loss}
\end{wraptable}

%% file: Discussion.tex
\vspace{-0.05in}
\section{Discussion} 
\vspace{-0.05in}
As shown above, CascadeXML substantially outperforms current state-of-the-art approaches both in terms of computational efficiency, and prediction performance. 
It achieves this while being simpler than XR-Transformer and more scalable than LightXML. Revisiting \autoref{tab:xmc-methods}, the extraction of intermediate [CLS] representations enables us to perform end-to-end multi-resolution training with resolution-specific attention maps from a single transformer model. 
From \autoref{tbl:results}, we show improvements over both LightXML and XR-Transformer, which are transformer-based models sharing feature representations across multiple label resolutions. We attribute this increased performance to CascadeXML's end-to-end multi-resolution training pipeline and property of adapted feature embeddings.
The presented architecture, therefore, combines the strength of different previous approaches, without inheriting
their limitations.

We postulated that the (meta-)classification tasks at different resolutions need to attend to different tokens in the input sequence, a property lacking in recent approaches \cite{Chang20, Jiang2021, zhang2021fast}. 
This intuition is corroborated by the fact that the attention maps learned by CascadeXML differ significantly between different resolutions. We provide graphs and supporting data in \autoref{appn:visual}. 

Obtaining label resolution specific feature embeddings in a pre-trained transformer model requires to make the compromise of extracting these at earlier transformer layers, where the token representations are not yet as refined as in the final layer \cite{xin-etal-2020-deebert}.
In contrast to the extreme classification task, which requires placing true labels exactly at the top-5 positions, the shortlisting task only requires all true meta-labels be recalled within the shortlist, which is much longer than five elements. 
We observe that the meta-classifiers are able to achieve high-recall (as shown \autoref{appn:visual}) at the shortlist length even with the less refined features. CascadeXML is, therefore, successful in efficiently leveraging intermediate transformer representations for the task of label shortlisting while keeping feature representations separate for each label resolution.

The structure of the model means that earlier layers in the transformer receive gradients from multiple resolutions, whereas the final layers' back propagation is limited to gradients from the extreme resolution (cf. \autoref{fig:diagram}).
This setup ensures that a portion of the representational capabilities is exclusively available for the extreme task and hence, the final transformer layers are increasingly more suitable for classification at the extreme resolution. 
At the same time, since the gradients from the finer tasks are also relevant for the coarser meta-label shortlisting task, so additional downstream task does not hurt the shortlisting performance at the intermediate transformer layers. 

%% file: Related.tex
\vspace{-0.05in}
\section{Other Related Works}
\vspace{-0.05in}
A large majority of the initial works in XMC have been focussed on learning the classifier (with fixed features), with one of the following class of methods : (i) one-vs-rest \cite{ Yen16, Babbar17, Yen17, Babbar19, schultheis2021speeding}, (ii) label trees \cite{Prabhu18, Khandagale19, Wydmuch18, Jasinska16}, (iii) decision trees \cite{Prabhu14, siblini2018craftml, majzoubi2020ldsm}, and (iv) label embedding based \cite{Bhatia15, Tagami17, Guo2019} classifiers. 
Beyond the above algorithmic categorization of approaches, computational considerations on scaling the training process via negative sampling \cite{Jain19}, and smart initialization \cite{fang2019fast, schultheis2021speeding} has also been studied. 
Furthermore, the statistical consequences of negative sampling \cite{Reddi2018} and missing labels \cite{jain2016extreme, qaraei2021convex,schultheis2022missing, schultheis2021unbiased, wydmuch2021propensity} have led to novel insights in designing unbiased loss functions.

With advances in deep learning, joint learning of features and classifier has been the focus of most recent approaches.
Building on the first steps for developing convolutional neural networks for text classification \cite{Kim14}, deep extreme classification was introduced in XML-CNN \cite{liu2017deep}. 
This work preceded the recent developments on employing BiLSTM and transformer encoders discussed in detail the Introduction section.
There has been a recent surge in works to tackle XMC for instances with short text inputs and those in which labels are endowed with textual descriptions \cite{Dahiya21,Dahiya21b,Mittal21,Mittal21b,Saini21}. 
Lately, the XMC setting has also been extended to predict unseen labels in zero-shot learning \cite{Gupta21,zhang2022metadata, xiong2021extreme}.

%% file: Conclusion.tex
\vspace{-0.08in}
\section{Conclusion}
\vspace{-0.08in}
In this paper, we introduced a novel paradigm for fine-tuning transformer models in XMC settings. 
In contrast to the existing methods, which consider transformer as a blackbox encoder, we leverage its multi-layered architecture to learn data representation corresponding to different resolutions in the HLT as well as fine-grained labels at the level of the extreme classifier. 
The proposed instantiation of our framework in the form of CascadeXML not only yields state-of-the-art prediction performance, but is also end-to-end trainable (without intermediate reclustering steps), simpler to implement, and fast on training and inference.
Beyond further research in extreme classification towards fully harnessing the representation capabilities of transformer encoders, we believe our approach can inspire future multi-resolution architectures in other domains as well which leverage label hierarchy.
\vspace{-0.08in}
\section{Acknowledgments}
\vspace{-0.08in}
The authors would like to thank Devaansh Gupta and Mohammadreza Qaraei for useful discussions. They also acknowledge the support of CSC – IT Center for Science, Finland, as well as the Aalto Science-IT project, for providing the required computational resources. This research is supported in part by Academy of Finland grants : Decision No. 348215 and 347707. 

%% file: AppendixA.tex
\appendix
\section{Appendix}

\subsection{Dataset Details \& Evaluation Metrics}

As stated earlier, the main application of Extreme Multi-label Text Classification is in e-commerce - product recommendation and dynamic search advertisement - and in document tagging, where the objective of an algorithm is to correctly recommend/advertise among the top-k slots. 
Thus, for evaluation of the methods, we use precision at $k$ (denoted by $P@k$), and its propensity scored variant (denoted by $PSP@k$) \cite{jain2016extreme}. 
These are standard and widely used metrics by the XMC community \cite{ExtremeRepository}.
For each test sample with observed ground truth label vector $y \in \{0,1\}^\numlabels$ and predicted vector $\hat{y} \in \mathbb{R}^\numlabels$, $P@k$ is given by :
{
\begin{align}
\mathrm{P@}k(\mathbf{y},\hat{\mathbf{y}}) \coloneqq &\frac{1}{k} \sum_{{\ell}\ \in\ \mathrm{top@}k(\hat{\mathbf{y}})} y_{\ell} \nonumber
\end{align}
where $\mathrm{top@}k(\hat{y})$ returns the $k$ largest indices of $\hat{y}$.
}

Since $\mathrm{P@}k$ treats all the labels equally, it doesn't reveal the performance of the model on tail labels. 
However, because of the long-tailed distribution in XMC datasets, one of the main challenges is to predict tail labels correctly, which may be more valuable and informative compared to head classes. A set of metrics that have been established in XMC to evaluate tail performance are propensity-scored version of precision. These $\mathrm{PSP}@k$ were introduced in \cite{jain2016extreme}, and use a weighting factor based on a propensity score $p_\ell$ to give more weight to tail labels:
\begin{align}
\mathrm{PSP@}k(\mathbf{y},\hat{\mathbf{y}}) \coloneqq &\frac{1}{k} \sum_{{\ell}\ \in\ \mathrm{top@}k(\hat{\mathbf{y}})} \frac{y_{\ell}}{p_\ell} \nonumber.
\end{align}
We use the empirical values for $p_\ell$ as proposed in \cite{jain2016extreme}.

\input{TableDatasets.tex}

\subsection{Model Details}


\input{TableHyperparameters.tex}

\subsubsection{Model Hyperparameters}
CascadeXML optimizes the training objective using Binary Cross Entropy loss as the loss function and AdamW \cite{AdamW} as the optimizer. We use different learning rates for the transformer encoder and the (meta-)label weight vectors as we need to train the weight vectors from random initialization in contrast to fine-tuning the transformer encoder. 
Specifically, the transformer encoder is fine-tuned at a learning rate of $10^{-4}$, while the weight vectors are trained at a learning rate of $10^{-3}$. The learning rate schedule consists of a constant learning rate for most of the iterations, with a cosine warm-up at the beginning and cosine annealing towards the end of the schedule. 
In multi-GPU setting, we use a batch size of 64 per GPU (256 total) across 4 GPUs. In single GPU setting, we still use a batch size of 256 by accumulating gradients for 4 iterations. The hyperparameter settings in detail have been mentioned in \autoref{tbl:hyperparams}.

\subsubsection{Ensemble Training Time}
As shown in \autoref{tbl:training_time}, CascadeXML achieves the lowest training time across all datasets except Amazon-3M using only 4 GPUs as compared to XR-Transformer which leverages 8 GPUs. Note that on Amazon-3M, XR-Transformer achieves a slightly lower training time. However, XR-Transformer uses 2$\times$ the number of GPUs and does not train the DNN model on full 3 million label resolution. XR-Transformer trains the DNN model using a classification training object comprising of only $2^{15}$ label clusters \cite[Table 7]{zhang2021fast} and then leverages XR-Linear \cite{yu2020pecos}, a linear solver, to scale up to 3M labels. 
On the other hand, CascadeXML trains an ensemble of 3 models to full resolution of 3 million labels in 30 hours using only 4 GPUs. 

\input{TableTrainingTime.tex}

\subsection{Leveraging Sparse Features}
\label{sec:Dismec}
As we are using BERT for the transformer backbone of our method, we have to truncate the input sequences to a limited number of tokens (see \autoref{tbl:hyperparams}).
This truncation results in loss of information. Thus, following XR-Transformer's lead we combine the features trained by CasadeXML with statistical information in the form of sparse tf-idf representation of the full input in an additional OVA classifier. The concatenated features are constructed as \cite{Jiang2021}:w
\begin{equation*}
    \encoder[cat] (\instance) = \left[\frac{\encoder[dnn](\instance)}{\|\encoder[dnn](\instance)\|} , \frac{\encoder[tf-idf](\instance)}{\|\encoder[tf-idf](\instance)\|} \right]
\end{equation*}

We use a version of DiSMEC \cite{Babbar17} instead of using XR-Linear \cite{yu2020pecos} - as done in XR-Transformer - as our external OVA classifier for $\encoder[cat]$. Even though XR-Linear achieves slightly better performance than DiSMEC across datasets (Table: 1), we find DiSMEC to be more resource efficient than XR-Linear. To quantify, DiSMEC runs successfully on 116GB RAM for all datasets, while XR-Linear requires close to 470GB RAM for Amazon-3M. 
Next, we discuss the application of DiSMEC over $\encoder[cat]$. 

\paragraph{DiSMEC} DiSMEC is a linear multilabel classifier that minimizes an $L_2$-regularized squared hinge-loss, followed by a pruning step to only keep the most important weights. Thus, the loss for a given weight matrix $\mathbf{W} = [\mathbf{w}_1, \ldots, \mathbf{w}_\numlabels]$ is given by 
\begin{equation}
    \mathcal{L}[\mathbf{W}] = \lambda \|\mathbf{W}\|_2^2 + \sum_{i=1}^{N} \sum_{j=1}^{\numlabels} \operatorname{max}\left(0, 1 - y_{ij} \langle \encoder[cat] (\instance_i), \mathbf{w}_j \rangle \right)^2.
\end{equation}
Crucially, from the point of view of the OVA classifiers, the input features $\encoder[cat] (\instance_i)$ are constant. This means that the task decomposes into independent optimization problems for each label, minimizing
\begin{equation}
    \mathcal{L}[\mathbf{w}_j] = \lambda \|\mathbf{w}_j\|_2^2 + \sum_{i=1}^{N} \operatorname{max}\left(0, 1 - y_{ij} \langle \encoder[cat] (\instance_i), \mathbf{w}_j \rangle \right)^2. \label{eq:dismec-loss}
\end{equation}
This allows for trivial parallelization of the task across CPU cores, and also means that the weights $\mathbf{w}_j$ can be pruned as soon as the sub-problem is solved. Consequently, there is no need to ever store the entire weight matrix, improving memory efficiency. 

The objective function \eqref{eq:dismec-loss} has a continuous derivative, and its Hessian is well defined everywhere except exactly at the decision boundary. 
Consequently, it can be minimized using a second-order Newton method. A discussion of this in the context of linear classification can be found in \cite{galli2021study}.

%% file: TableDatasets.tex
\begin{table*}[h]
\centering
\small
\begin{adjustbox}{width=0.9\textwidth,center}
\scalebox{0.95}{
\begin{tabular}{c|c|c|c|c|c|c}
\toprule
\textbf{Datasets} & $\textbf{d}_{\textrm{tf-idf}}$ & \textbf{\# Labels} & \textbf{\# Training} & \textbf{\# Test} & \textbf{ALpP} & \textbf{APpL}\\
\midrule
\textbf{Wiki10-31K} & 101,938  & 30,938 & 14,146 & 6,616 & 18.64 & 8.52 \\
\textbf{AmazonCat-13K} & 203,882  & 13,330 & 1,186,239 & 306,782 & 5.04 & 448.57 \\
\textbf{Wiki-500K} & 2,381,304 & 501,070 & 1,779,881 & 769,421 & 4.75 & 16.86 \\
\textbf{Amazon-670K} & 135,909  & 670,091 & 490,449 & 153,025 & 5.45 & 3.99 \\
\textbf{Amazon-3M} & 337,067 & 2,812,281 & 1,717,899 & 742,507 & 36.04 & 22.02 \\
\midrule
\end{tabular}
}
\end{adjustbox}
\vspace{-3mm}
\caption{Dataset Statistics. APpL denotes the average data points per label, ALpP the average number of labels per point. For a fair comparison with other baselines, we download these five publicly available benchmark datasets from \href{https://github.com/yourh/AttentionXML}{https://github.com/yourh/AttentionXML}.
}\label{tbl:datasets}
\end{table*}

%% file: TableHyperparameters.tex
\begin{table*}[h]
\centering
\small
\begin{adjustbox}{width=\textwidth,center}
\scalebox{0.95}{
\begin{tabular}{c|c|c|c|c|c}
\toprule
\textbf{Datasets} & Transformer Layer : Label Resolutions & Shortlist Size & Ep & $N_x$ & Dropouts\\
\midrule
\textbf{Wiki10-31K} & {\{9, 10\} : $2^9$ | 11 : $2^{12}$ | 12 : 30938}  & {$2^9$, $2^9$, \textasciitilde $2^9$} & 15 & 256 & {0.3, 0.3, 0.4} \\
\textbf{AmazonCat-13K} & {\{7, 8\} : $2^8$ | 10 : $2^{11}$ | 12:13330} & {$2^8$, $2^8$, \textasciitilde $2^8$}  & 6 & 256 & {0.2, 0.3, 0.4} \\
\textbf{Wiki-500K} & {\{5, 6\} : $2^{10}$ | 8 : $2^{13}$ | 10 : $2^{16}$ | 12 : 501070} & {$2^{10}$, $2^{10}$, $2^{11}$, \textasciitilde $2^{12}$}  & 12 & 128 & {0.2, 0.25, 0.35, 0.5} \\
\textbf{Amazon-670K} & {\{5, 6\} : $2^{10}$ | 8 : $2^{13}$ | 10 : $2^{16}$ | 12 : 670091} & {$2^{10}$, $2^{10}$, $2^{11}$, \textasciitilde $2^{12}$}  & 15 & 128 & {0.2, 0.25, 0.4, 0.5} \\
\textbf{Amazon-3M} & {\{5, 6\} : $2^{10}$ | 8 : $2^{13}$ | 10 : $2^{16}$ | 12 : 2812281} & {$2^{10}$, $2^{10}$, $2^{11}$, \textasciitilde $2^{12}$}  & 15 & 128 & {0.2, 0.25, 0.3, 0.5}\\
\midrule
\end{tabular}
}
\end{adjustbox}
\caption{Hyperparameters for CascadeXML. Ep denotes the total number of epochs needed to fine-tune model over the dataset. $N_x$ is the number of text tokens input to the model after truncation. Transformer layers put in brackets next to the label resolution imply that a concatenation of the [CLS] token of the respective layers has been used for label shortlisting at that resolution.}
\label{tbl:hyperparams}
\end{table*}

%% file: TableTrainingTime.tex
\begin{table}[h]
\begin{adjustbox}{width=\textwidth,center}
\begin{tabular}{c|c|c|c|c|c}
\toprule
Dataset & AttentionXML-3 & X-Transformer-9 & LightXML-3 & XR-Transformer-3 & CascadeXML-3 \\
\midrule
Wiki10-31K & 1.5 & 14.1 & 26.9 & 1.5 & \textbf{0.4} \\
AmazonCat-13K & 24.3 & 147.6 & 310.6 & 13.2 & \textbf{9.8}\\
Wiki-500K & 37.6 & 557.1 & 271.3 & 38.0 & \textbf{21.6}\\
Amazon-670K   & 24.2 & 514.8 & 159.0 & 10.5 & \textbf{9.0} \\
Amazon-3M & 54.8 & 542.0 & - & \textbf{29.3} & 30.0 \\ 
\bottomrule
\end{tabular}
\end{adjustbox}
\vspace{1pt}
\caption{Time taken to train the ensembles of the respective models. Training time AttentionXML, X-Transformer and XR-Transformer have been reported using 8 NVidia V100 GPUs. Training time for LightXML is clocked using 1 GPU and that of CascadeXML is clocked using 4 GPUs.}   
\label{tbl:training_time}
\end{table}

%% file: AppendixB.tex
\section{Visualizations and Analysis}
\label{appn:visual}
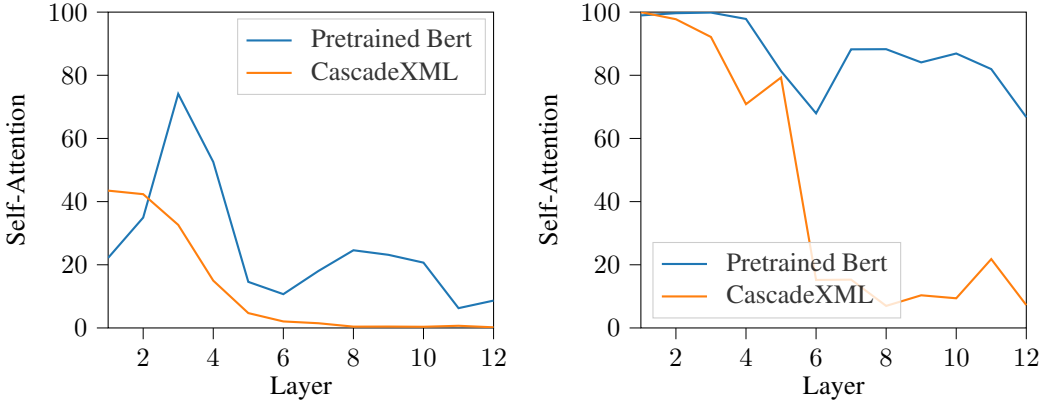
\begin{figure}[h]
  \input{FigureSelf-attention-mean.tex}
  \input{FigureSelf-attention-max.tex}
  \caption{Average (left) and maximum (right) self-attention of the [CLS] token at a given layer to the [CLS] token in 
  the preceding layer. The embedding for the [CLS] token is almost exclusively assembled from the embeddings of the other
  tokens in the later layers much more strongly in CascadeXML than in pretrained BERT.
  \label{fig:self-attn}
  }
\end{figure}

In this section we provide visualizations and additional data that corroborate our interpretation that different attention- and feature maps are needed for classification at different granularities of the label tree. 
In \autoref{fig:self-attn}, the attention of the [CLS] token to itself in the previous layer is visualized. 

If this self-attention were large, then $\encoder[CLS]^{(t)}$ would be mostly a function of $\encoder[CLS]^{(t-1)}$. In
such a case, $\encoder[CLS]^{(t-1)}$ would contain less information than $\encoder[CLS]^{(t)}$ (data-processing
inequality), but the meta-labels $\resolution{t-1}(\labelvec)$ at level $t-1$ contain strictly less information than
$\resolution{t}(\labelvec)$. Thus, is the [CLS] token embedding had strong feed-forward characteristics, $\encoder[CLS]^{(t-1)}$
would have to contain all the information also about the extreme-level labels, and thus have limited representation capacity
for the level-t task. 

Luckily, \autoref{fig:self-attn} is a sanity-check that shows that this is not the case. 
Starting from layer $6$, where the first meta-task is placed, the [CLS] token is almost entirely re-assembled at each layer from the embeddings of the other tokens -- much more strongly than in a pretrained BERT.
This allows each layer to extract the information best suited for classification at the given hierarchy level.

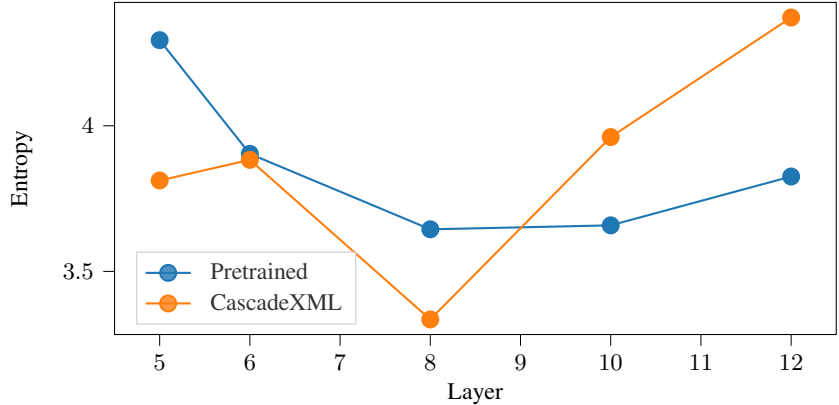
\begin{figure}[h]
\centering
\input{FigureEntropy.tex}
\caption{Entropy of the distribution of attention to the input tokens. A large value indicates that attention is given to many different tokens, whereas a smaller value means that few tokens receive most of the attention.\label{fig:entropy}}
\end{figure}

We can also detect some qualitative differences in the attention maps at different resolutions: The entropy, i.e. how much the attention is concentrated or spread across different tokens, changes significantly between levels. 
This is not an artifact of the pretrained BERT model, but appears to be learned during fine-tuning. 

\begin{figure}[h]
\centering
\input{FigurePwcca.tex}
\caption{PWCCA similarity between [CLS] token representations at different levels for CascadeXML trained on Amazon-670K dataset. For this dataset, we place the weight vectors of different label resolutions at layers 6, 8, 10 and 12.}
\label{pwcca}
\end{figure}
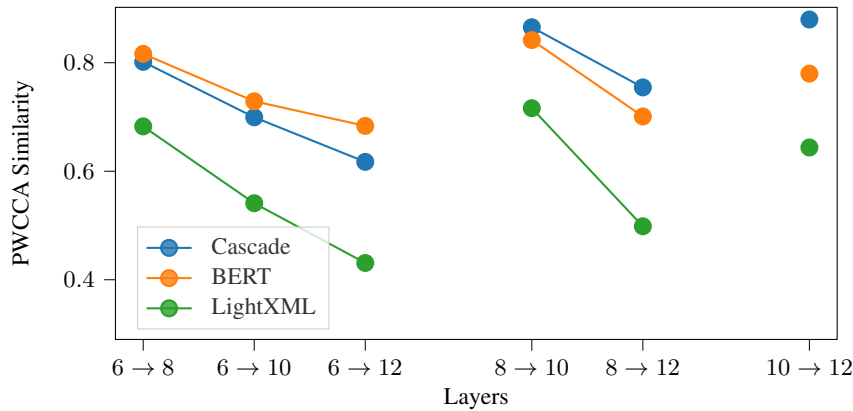

In \autoref{pwcca}, we analyse the flow of information and visualize how much processing is happening in a given layer with respect to the [CLS] token.
We rely on projection weighted canonical correlation analysis \cite{PWCCA} for this task. This allows to compare the representations
at different layers in a way that is invariant to any linear transformations. 

We primarily make two observations from \autoref{pwcca}. First, we observe that [CLS] token embeddings of layers 6, 8, 10 and 12 are more closely related in CascadeXML than in LightXML. This is expected as LigthXML only trains the (meta-)label weight vectors using the [CLS] token embeddings of the bottom layers. 
Because the multi-resolution training objectives differ only in granularity, many features required to distinguish coarse meta-labels are useful in determining finer meta-labels as well. Hence, the similarity between consecutive representations is expected to be strong. 
On the other hand, when only looking at CascadeXML's points (in blue) in \autoref{pwcca}, we observe that the tasks in the first meta-classifier and the extreme classifier are substantially different. 
This implies both training objectives require different representations that cannot be provided by a LightXML-/XR-Transformer-style model which use same attention maps (and hence, same [CLS] token embeddings) for all resolutions. 

\begin{table}[h]
\centering
\caption{Recall of the shortlisting tasks.}
\label{tab:recall}
\pgfplotstabletypeset[ every head row/.style={before row={\toprule}, after row=\midrule,},
    every last row/.style={after row=\bottomrule}, columns/Dataset/.style={string type, column type=l},]{
{Dataset}     {Level 1}     {Level 2}     {Level 3}
Amazon-670K   98.29         91.93         83.10
Wiki-500L     99.54         96.53         93.12
}
\end{table}

The ability to use earlier layers' [CLS] token representation for the 
meta-task crucially depends on the fact that these representations are
still sufficient for achieving high recall in the shortlisting task. 
As shown in \autoref{tab:recall}, the shortlisting achieves very good recall rates. In particular, the very first shortlist, with the \enquote{weakest} features, achieves almost perfect recall.

%% file: FigureSelf-attention-mean.tex
\begin{tikzpicture}

\definecolor{color0}{rgb}{0.12156862745098,0.466666666666667,0.705882352941177}
\definecolor{color1}{rgb}{1,0.498039215686275,0.0549019607843137}

\begin{axis}[
legend cell align={left},
legend style={fill opacity=0.8, draw opacity=1, text opacity=1, draw=white!80!black},
tick align=outside,
tick pos=left,
width=0.48\textwidth,
x grid style={white!69.0196078431373!black},
xlabel={Layer},
xmin=1, xmax=12,
xtick style={color=black},
y grid style={white!69.0196078431373!black},
ylabel={Self-Attention},
ymin=0, ymax=100,
ytick style={color=black}
]
\addplot [thick, color0]
table {%
1 22.1846046447754
2 34.9391174316406
3 74.1045227050781
4 52.5643005371094
5 14.5888805389404
6 10.6976594924927
7 18.0272808074951
8 24.5958518981934
9 23.1527080535889
10 20.6775817871094
11 6.2662615776062
12 8.66202926635742
};
\addlegendentry{Pretrained Bert}
\addplot [thick, color1]
table {%
1 43.4768562316895
2 42.3561630249023
3 32.6342010498047
4 14.981987953186
5 4.70381355285645
6 2.06087756156921
7 1.49206030368805
8 0.414554864168167
9 0.438662469387054
10 0.37375408411026
11 0.683020830154419
12 0.200836032629013
};
\addlegendentry{CascadeXML}
\end{axis}

\end{tikzpicture}

%% file: FigureSelf-attention-max.tex
\begin{tikzpicture}

\definecolor{color0}{rgb}{0.12156862745098,0.466666666666667,0.705882352941177}
\definecolor{color1}{rgb}{1,0.498039215686275,0.0549019607843137}

\begin{axis}[
legend cell align={left},
legend style={fill opacity=0.8, draw opacity=1, text opacity=1, at={(0.03,0.03)}, anchor=south west, draw=white!80!black},
tick align=outside,
tick pos=left,
width=0.48\textwidth,
x grid style={white!69.0196078431373!black},
xlabel={Layer},
xmin=1, xmax=12,
xtick style={color=black},
y grid style={white!69.0196078431373!black},
ylabel={Self-Attention},
ymin=0, ymax=100,
ytick style={color=black}
]
\addplot [thick, color0]
table {%
1 98.9561233520508
2 99.6614532470703
3 99.8562774658203
4 97.8362655639648
5 81.3175964355469
6 67.9516220092773
7 88.2145080566406
8 88.262939453125
9 84.0828018188477
10 86.8965835571289
11 81.9540786743164
12 66.7397689819336
};
\addlegendentry{Pretrained Bert}
\addplot [thick, color1]
table {%
1 99.9935989379883
2 97.7469863891602
3 92.1273880004883
4 70.8547286987305
5 79.2921905517578
6 15.1860818862915
7 15.2975168228149
8 6.97629356384277
9 10.3393659591675
10 9.39135932922363
11 21.797981262207
12 7.22851896286011
};
\addlegendentry{CascadeXML}
\end{axis}

\end{tikzpicture}

%% file: FigureEntropy.tex
\begin{tikzpicture}

\definecolor{color0}{rgb}{0.12156862745098,0.466666666666667,0.705882352941177}
\definecolor{color1}{rgb}{1,0.498039215686275,0.0549019607843137}

\begin{axis}[
legend cell align={left},
legend style={fill opacity=0.8, draw opacity=1, text opacity=1, at={(0.03,0.03)}, anchor=south west, draw=white!80!black},
tick align=outside,
tick pos=left,
width=0.8\textwidth,
height=6cm,
x grid style={white!69.0196078431373!black},
xlabel={Layer},
xmin=4.5, xmax=12.5,
xtick style={color=black},
y grid style={white!69.0196078431373!black},
ylabel={Entropy},
ymin=3.28325306177139, ymax=4.42358006238937,
ytick style={color=black},
font=\small
]
\addplot [thick, color0, mark=*, mark size=3, mark options={solid}]
table {%
5 4.294029712677
6 3.90360045433044
8 3.64449214935303
10 3.6581289768219
12 3.82576179504395
};
\addlegendentry{Pretrained}
\addplot [thick, color1, mark=*, mark size=3, mark options={solid}]
table {%
5 3.81216597557068
6 3.8829710483551
8 3.33508610725403
10 3.96130084991455
12 4.37174701690674
};
\addlegendentry{CascadeXML}
\end{axis}

\end{tikzpicture}

%% file: FigurePwcca.tex
\begin{tikzpicture}

\definecolor{color0}{rgb}{0.12156862745098,0.466666666666667,0.705882352941177}
\definecolor{color1}{rgb}{1,0.498039215686275,0.0549019607843137}
\definecolor{color2}{rgb}{0.172549019607843,0.627450980392157,0.172549019607843}

\begin{axis}[
legend cell align={left},
legend style={fill opacity=0.8, draw opacity=1, text opacity=1, at={(0.03,0.03)}, anchor=south west, draw=white!80!black},
tick align=outside,
tick pos=left,
width=0.8\textwidth,
height=6cm,
x grid style={white!69.0196078431373!black},
xlabel={Layers},
xmin=-0.25, xmax=6.25,
xtick style={color=black},
xtick={0,1,2,3.5,4.5,6},
xticklabels={{$6\rightarrow8$},{$6\rightarrow10$},{$6\rightarrow12$},{$8\rightarrow10$},{$8\rightarrow12$},{$10\rightarrow12$}},
y grid style={white!69.0196078431373!black},
ylabel={PWCCA Similarity},
ymin=0.29, ymax=0.9020868055,
ytick style={color=black},
font=\small
]
\addplot [thick, color0, mark=*, mark size=3, mark options={solid}]
table {%
0 0.80163402
1 0.69942173
2 0.61730222
};
\addlegendentry{Cascade}
\addplot [thick, color1, mark=*, mark size=3, mark options={solid}]
table {%
0 0.81634159
1 0.72897326
2 0.68357666
};
\addlegendentry{BERT}
\addplot [thick, color2, mark=*, mark size=3, mark options={solid}]
table {%
0 0.68264861
1 0.54105831
2 0.43091347
};
\addlegendentry{LightXML}
\addplot [thick, color0, mark=*, mark size=3, mark options={solid}, forget plot]
table {%
3.5 0.86538677
4.5 0.75451613
};
\addplot [thick, color1, mark=*, mark size=3, mark options={solid}, forget plot]
table {%
3.5 0.84179357
4.5 0.70097982
};
\addplot [thick, color2, mark=*, mark size=3, mark options={solid}, forget plot]
table {%
3.5 0.7161774
4.5 0.49855842
};
\addplot [thick, color0, mark=*, mark size=3, mark options={solid}, forget plot]
table {%
6 0.87964998
};
\addplot [thick, color1, mark=*, mark size=3, mark options={solid}, forget plot]
table {%
6 0.77992789
};
\addplot [thick, color2, mark=*, mark size=3, mark options={solid}, forget plot]
table {%
6 0.64379361
};
\end{axis}

\end{tikzpicture}